\documentclass[10pt,twocolumn,letterpaper]{article}

\usepackage[pagenumbers]{paperstyle}

\newcommand{\method}{DuoEdit}
\usepackage{multirow}
\usepackage{colortbl}
\usepackage{pifont} %
\usepackage{placeins} %
\usepackage{microtype}

\definecolor{linkblue}{rgb}{0.21,0.49,0.74}
\usepackage[breaklinks,colorlinks,allcolors=linkblue]{hyperref}
\hypersetup{
    pdftitle={DuoEdit: A Unified Local Video Editing Model Learning Only Two Primitive Operators},
    pdfauthor={Jiayi Wu, Haoming Cai, Cornelia Fermuller, Yezhou Yang, Furong Huang, Christopher Metzler, Yiannis Aloimonos}
}

\title{DuoEdit: A Unified Local Video Editing Model Learning Only Two Primitive Operators}

\author{%
Jiayi Wu$^{1,*}$ \quad Haoming Cai$^{1,*}$ \quad Cornelia Fermuller$^{1}$ \quad Yezhou Yang$^{2}$\\
Furong Huang$^{1}$ \quad Christopher Metzler$^{1}$ \quad Yiannis Aloimonos$^{1}$\\[0.4em]
{\normalsize $^{1}$University of Maryland, College Park \qquad $^{2}$Arizona State University}\\[0.2em]
{\normalsize Project Page: \url{https://jiayi-wu-leo.github.io/duoedit/}}
}

\makeatletter
\def\ps@duoeditfirstpage{%
  \let\@mkboth\@gobbletwo
  \let\@oddhead\@empty
  \let\@evenhead\@empty
  \def\@oddfoot{%
    \normalfont
    \rlap{\footnotesize $^{*}$Equal contribution.}%
    \hfil\thepage\hfil
  }%
  \let\@evenfoot\@oddfoot
}
\makeatother

\makeatletter
\patchcmd{\@maketitle}{\vspace*{24pt}}{\vspace*{6pt}}{}{%
    \PackageError{duoedit}{Could not adjust title spacing}{}
}
\g@addto@macro\@maketitle{%
    \noindent
\begin{minipage}{\textwidth}
    \centering
    \includegraphics[width=\linewidth]{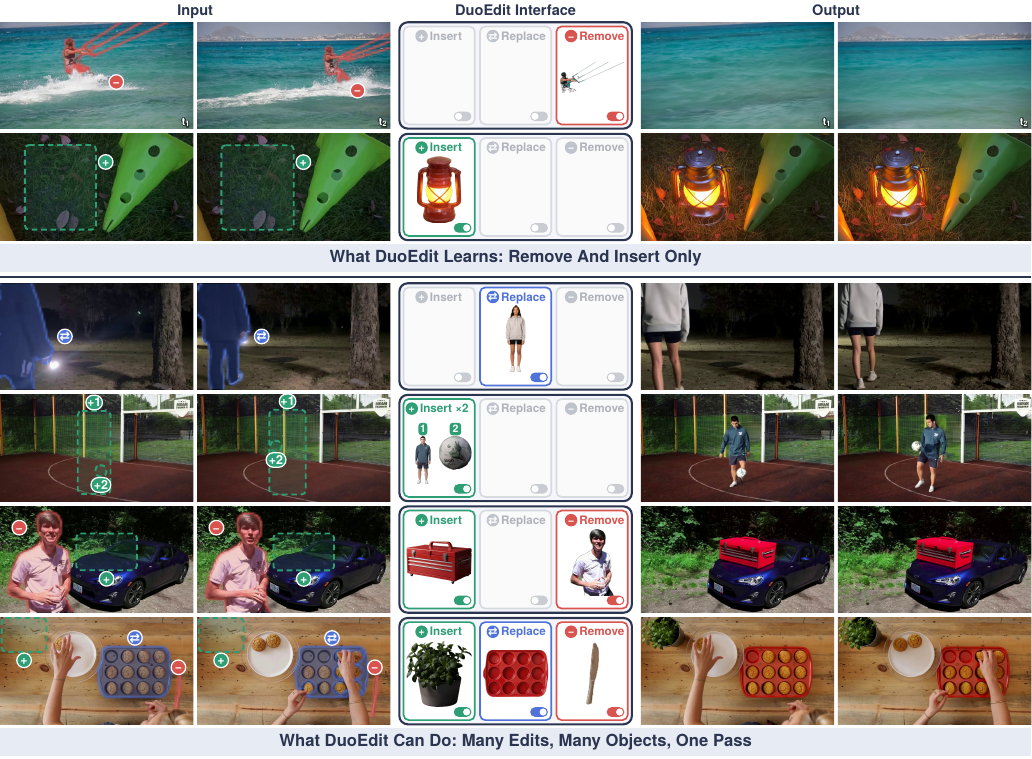}
    \par
    \captionsetup{hypcap=false,skip=4pt}
    \captionof{figure}{\textbf{Unified local video editing by learning only two primitive operators.}
    \method{} is trained only on removal and insertion (top two rows).
    It composes multiple operations across multiple subjects in one end-to-end inference run, jointly generating the edited video.
    The remaining rows show zero-shot compositions, including replacement, human--object interaction, insertion with removal, and mixed operations, without additional task-specific adaptation.
    The DuoEdit interface between the input and edited frames of each example enables the requested operations and associates each with an unaligned appearance reference or the object to be removed.
    Two aligned input and edited frames are shown for each example.}
    \label{fig:teaser}
\end{minipage}
    \par\vspace{4pt}%
}
\makeatother

\begin{document}
\raggedbottom
\maketitle
\thispagestyle{duoeditfirstpage}
\begingroup
\setlength{\parskip}{1.5pt plus 0.5pt minus 0.25pt}
\setlength{\dbltextfloatsep}{26pt plus 2pt minus 4pt}
\begin{abstract}
Local video editing requires fine-grained subject control while harmonizing associated scene effects.
These demands become more challenging when different edits involve multiple interacting subjects, yet high-quality paired training videos for such compositional edits are difficult to obtain.
We show that compositional editing does not require compositional training data: it can emerge from learning only two primitive operators, removal and insertion, on simple paired videos.
Based on this philosophy, we present \method{}, a unified local video editing model in which the two primitives are controlled by independent conditions.
Because these conditions are independent, they can be activated together, so replacement and mixed multi-subject edits are performed zero-shot in a single inference run.
Additionally, to accept unaligned appearance references as persistent context throughout diffusion, \method{} decouples appearance from spatial guidance and learns to route each reference to its target box trajectory using shared subject IDs, enabling precise spatial control over each subject.
Experimental results demonstrate that \method{} matches or surpasses state-of-the-art performance on removal and insertion, while producing more reliable zero-shot compositional edits than mainstream unified video editing models and sequential editing baselines.
\end{abstract}
    
\section{Introduction}
\label{sec:intro}

Local video editing aims to modify selected subjects in an existing video while preserving unrelated content and maintaining scene coherence~\cite{zhou2023propainter,gu2024videoswap}.
It requires fine-grained control over individual subjects together with appropriate changes to their shadows, reflections, and occlusions~\cite{fu2026effecterase,miao2025rose}.
As illustrated in Fig.~\ref{fig:teaser}, requests range from applying the same operation to several subjects to combining removal, insertion, and replacement across interacting subjects.

Realizing these compositions raises three challenges.
First, models specialized for individual operations~\cite{miao2025rose,gao2026pisco} can be chained through multiple inference runs, but human--human and human--object interactions remain difficult to synthesize coherently through independent edits.
This motivates joint generation of the requested changes in a single inference run.
Second, some insertion pipelines require foreground sequences or keyframes aligned with the source video~\cite{gao2026pisco,chen2025contextflow}, with subject pose, scale, and placement largely determined in advance.
Preparing such inputs becomes demanding for multiple moving and interacting subjects, motivating an interface that accepts unaligned appearance references and separate spatial guidance.
Third, high-quality paired videos for compositional editing are inherently difficult to obtain~\cite{long2026coinve}, particularly when multiple operations involve interacting subjects and coordinated scene effects.

We present \method{}, a unified local video editing model that learns only two primitive operators: removal and insertion.
Erasure masks specify removal, while appearance references and bounding-box trajectories specify insertion.
These conditions can be independently specified and jointly activated: replacement combines erasure of an existing subject with insertion of a new one, while mixed requests assemble the controls across multiple subjects.
\method{} executes all requested operations in a single end-to-end inference run conditioned on the complete source video, jointly considering harmonization among edited subjects and with the surrounding scene.

\method{} decouples appearance references from spatial guidance, allowing users to specify subject appearance and target placement independently without constructing aligned foreground layers.
Appearance references need not be aligned with the target video and remain available as global appearance sources throughout diffusion.
Shared identity cues guide internal spatial--appearance routing, associating each reference with its intended bounding-box trajectory.
Text complements these controls with information about object dynamics, human actions, facial expressions, and scene effects that spatial conditions alone cannot readily describe.

We train both primitives within this shared interface using only an object-removal video dataset~\cite{fu2026effecterase}.
The original source--target pairs supervise removal, while reversing their direction provides insertion supervision; appearance references and trajectories are derived from the same videos.
Each training example contains only insertion or only removal, with no additional paired training videos for replacement or mixed-operation editing.
Quantitative and qualitative evaluations demonstrate strong removal and insertion performance, as well as effective zero-shot replacement and mixed-operation editing across multiple subjects.

Our contributions are threefold:
\begin{itemize}
    \item A unified local video editing model that learns only two primitive operators and performs multi-subject, multi-operation compositions in one inference run, jointly considering subject interactions and scene harmonization.
    \item A fine-grained control interface that keeps unaligned references globally available throughout diffusion and associates them with their intended box trajectories through internal spatial--appearance routing.
    \item A bidirectional training strategy that learns both primitives from the same object-removal video pairs, requiring no additional paired training videos for replacement or mixed-operation editing.
\end{itemize}

\begin{figure*}[t]
    \centering
    \includegraphics[width=\textwidth]{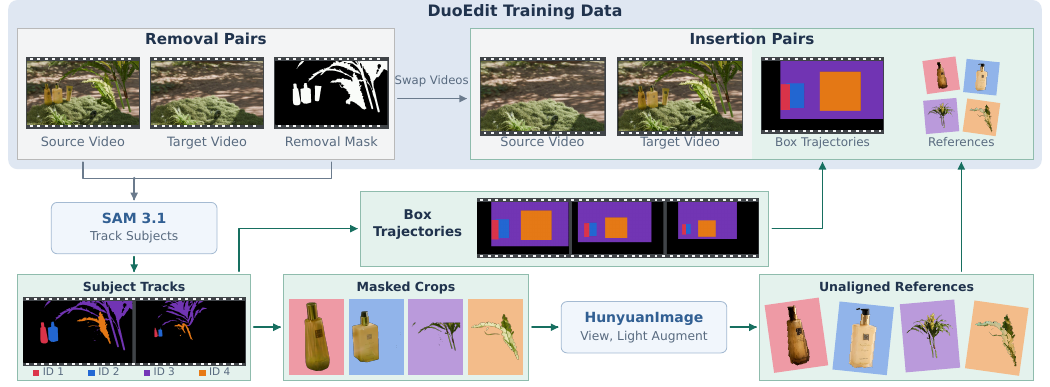}
    \caption{\textbf{Constructing bidirectional training data.}
    Removal samples provide paired videos with and without the subjects and their associated scene effects, together with removal masks, and are used directly for removal training.
    We reverse the video pairs for insertion.
    SAM~3.1 separates and tracks instances, yielding box trajectories with subject IDs and masked appearance crops.
    Image editing augments the crops into unaligned appearance references.
    The trajectories and references share IDs and complete the insertion conditions.
    Both directions train \method{} from the same video pairs.}
    \label{fig:data_pipeline}
\end{figure*}

\begin{figure*}[t]
    \centering
    \includegraphics[width=\textwidth]{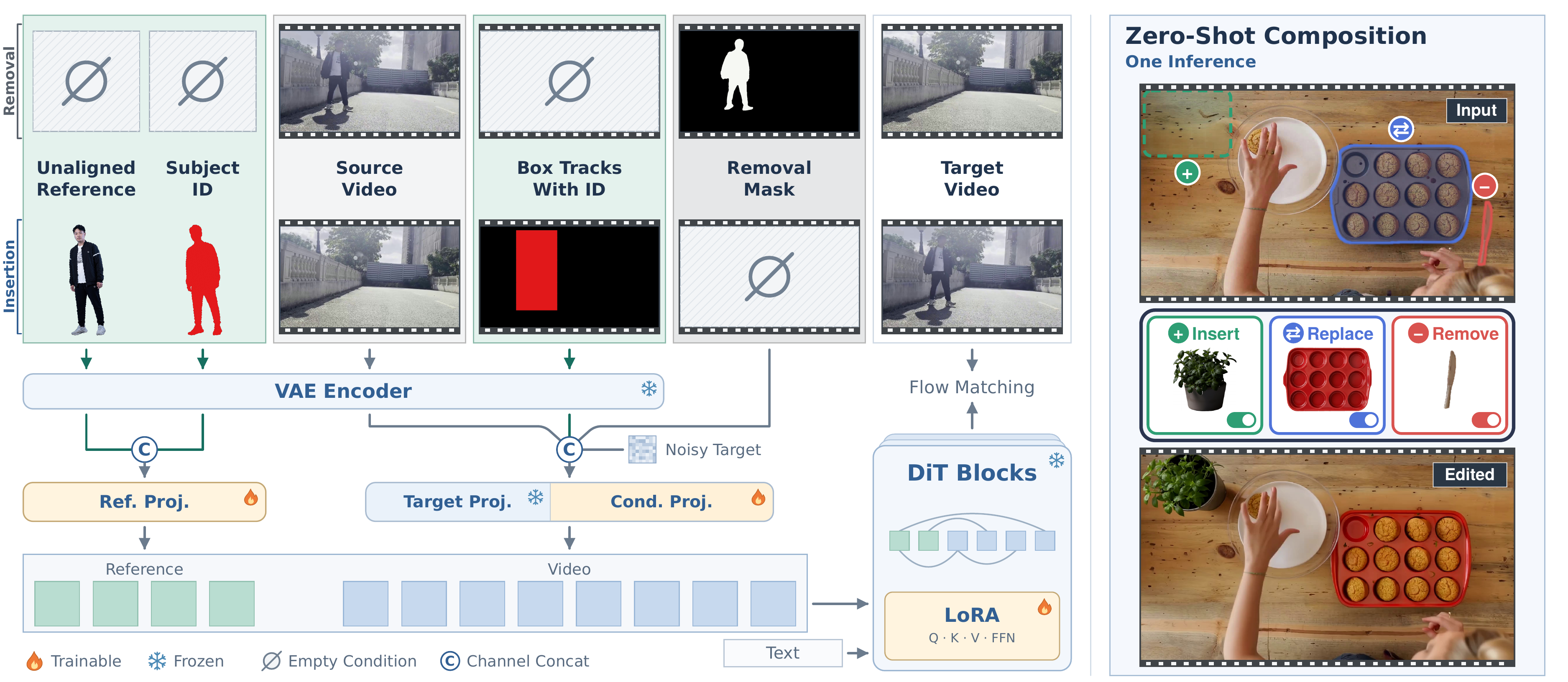}
    \caption{\textbf{Learning two primitives and composing edits at inference.}
    DuoEdit is trained only on removal and insertion.
    Subject IDs associate unaligned appearance references with their corresponding box trajectories.
    The two primitives use fully independent conditioning channels, enabling zero-shot composition of multiple operations across multiple subjects in a single inference run, without composition-specific training.}
    \label{fig:model_overview}
\end{figure*}

\section{Related Work}
\label{sec:related}

\subsection{Video Object Removal and Associated Effects}
Video inpainting provides a basis for object removal.
ProPainter~\cite{zhou2023propainter} combines propagation with a sparse video Transformer, while VideoPainter~\cite{bian2025videopainter} uses a context encoder to condition video diffusion on the visible background.
Removing an object can also require changing regions outside its silhouette.
ROSE~\cite{miao2025rose} uses synthetic paired videos and difference-mask supervision to localize object-induced effects, while UnderEraser~\cite{liu2026undereraser} models object--effect relations to locate and remove these effects.
DualEraser~\cite{chen2026dualeraser} combines text--mask guidance with separate localization and preservation components.
VOID~\cite{motamed2026void} further considers how object deletion changes downstream physical interactions.
Closely related to our training setup, EffectErase~\cite{fu2026effecterase} introduces VOR, a dataset of paired videos with and without target objects and their effects, and jointly learns removal and its inverse insertion task to improve effect erasing.
These methods primarily focus on video object removal.
\method{} reuses paired videos from the removal dataset in both directions to learn insertion and removal as controllable primitives, enabling general multi-subject local video editing.

\subsection{Reference-Guided Video Object Insertion and Replacement}
Reference-guided video editing introduces a specified object appearance while maintaining the surrounding source content.
PISCO~\cite{gao2026pisco} conditions object insertion on the source video and instance cutouts provided at sparse keyframes.
GenCompositor~\cite{yang2026gencompositor} edits a background video using dynamic foreground video references under trajectory and scale control, accommodating unaligned foreground and background layouts; it also reports removal and replacement extensions.
VideoSwap~\cite{gu2024videoswap} uses semantic point correspondences for subject replacement, preserving source motion while allowing changes in object shape.
ContextFlow~\cite{chen2025contextflow} supports insertion, swapping, and deletion by propagating an edited first frame with guidance from inversion and reconstruction of the source video.
Requiring foreground sequences, keyframe conditions, or point correspondences adds preparation effort when combining multiple edits.
\method{} separates appearance references from box trajectories and combines them with erasure masks, allowing edits to be composed without constructing foreground sequences or pre-editing source frames.

\subsection{Unified Video Editing Models}
Unified video editing methods support multiple transformations of an existing video within a common framework.
VACE~\cite{jiang2025vace} uses video inputs, masks, and optional reference images for editing, while OmniV2V~\cite{liang2025omniv2v} combines source-video conditions with visual references and editing instructions.
UniVideo~\cite{wei2025univideo} supports reference-guided object editing through multimodal instructions, and Omni-Video~2~\cite{yang2026omnivideo2} interprets the source video and editing instruction to guide the desired modification.
Bernini~\cite{liu2026bernini} introduces latent semantic planning, combining target semantic representations predicted by an MLLM with source-video features to guide video editing.
UniVideo also demonstrates unseen task compositions, including deleting one object while replacing another~\cite{wei2025univideo}.
These models unify diverse tasks, including local and global video editing, but balancing this breadth with fine-grained spatial control and reference appearance consistency remains challenging, especially when spatial intent is conveyed through text instructions~\cite{jiang2025vace,liang2025omniv2v}.
\method{} focuses on unified local video editing, binding each appearance reference to its corresponding box trajectory through internal spatial--appearance routing to provide fine-grained spatial and appearance control for multi-subject edit compositions.

\section{Method}
\label{sec:method}

\subsection{Problem Formulation: Unified Local Video Editing}
\label{sec:formulation}
\label{sec:primitives}
\label{sec:composition}

Given a source video $S$, unified local video editing seeks an output $\widehat V$ that satisfies subject-level edit requests while preserving unrelated scene content.
We specify the requested changes with an erasure mask $M$, bounding-box trajectories $B$, appearance references $R$, and a text instruction $p$:
\begin{equation}
    \widehat V=G_\theta(S,M,B,R,p).
    \label{eq:editor}
\end{equation}
Text complements the spatial and appearance conditions with details about motion, expressions, and scene effects.

The two primitives use orthogonal controls at the input-interface level: $M$ specifies erasure, while $(B,R)$ specifies synthesis.
These signals can be specified independently and activated jointly.
Replacement combines the old subject's erasure mask with the new subject's appearance and box trajectory.
The same rule extends across subjects to mixed-operation editing requests, which the model executes jointly without composition-specific training (\cref{fig:model_overview}).
All edits share the complete source context and a single denoising process.

\subsection{Constructing Bidirectional Supervision from Removal Data}
\label{sec:bidirectional_data}

We construct bidirectional supervision from object-removal data in VOR~\cite{fu2026effecterase}.
Each example contains an object-present video $V^{+}$, its object-absent counterpart $V^{-}$, and a removal mask provided by the dataset.
The video pair also captures corresponding changes in object-related effects.
As illustrated in \cref{fig:data_pipeline}, we use SAM~3.1~\cite{carion2025sam3} to track individual target subjects with distinct IDs and extract their appearance crops.
We augment these crops with HunyuanImage-3.0-Instruct-Distil~\cite{cao2025hunyuanimage,tencent2026hunyuanimageinstructdistil}, varying lighting and viewpoint and completing occluded regions to obtain unaligned appearance references.
A vision-language model (VLM) additionally generates editing instructions to complement these visual conditions.

Reversing the source $S$ and target $Y$ gives the two training directions:
\begin{equation}
    (S,Y)=
    \begin{cases}
        (V^{+},V^{-}), & \text{removal},\\
        (V^{-},V^{+}), & \text{insertion}.
    \end{cases}
    \label{eq:paired_tasks}
\end{equation}
Removal uses the provided mask with empty synthesis conditions.
Insertion uses the box trajectories and augmented appearance references with an empty erasure mask.
Both primitives are therefore learned from the same paired videos, without collecting additional video pairs.

\begin{figure*}[!t]
\centering
\includegraphics[width=\textwidth,height=0.88\textheight,keepaspectratio]{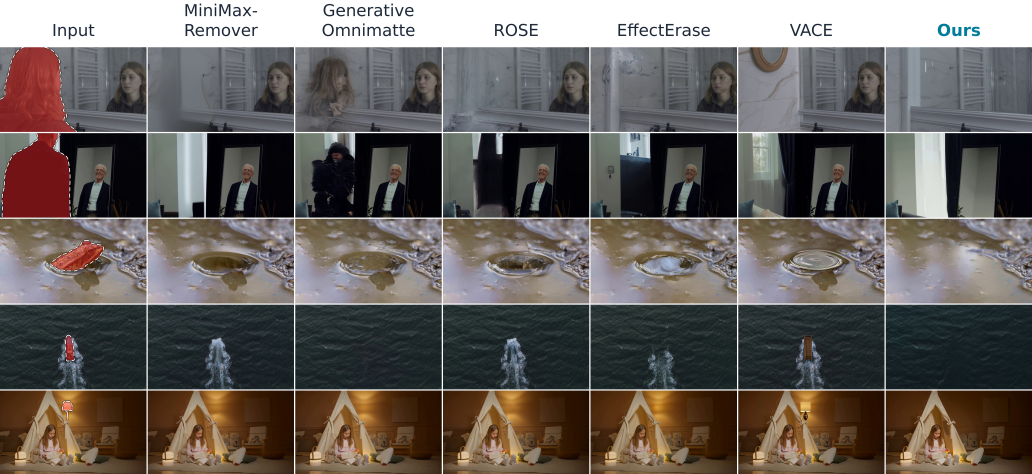}
\caption{\textbf{Qualitative comparison for video object removal on in-the-wild videos (VOR-Wild).} Each row is one video shown at a single frame; columns are methods, and the input shows the object mask in red. Ours removes the target object together with the effects it causes, such as mirror reflections (top two rows) and water ripples and wakes, while the baselines leave these effects behind or produce artifacts. \href{https://jiayi-wu-leo.github.io/duoedit/}{\textcolor{blue}{\textbf{More examples and full videos on the project page.}}}}
\label{fig:removal_qualitative}
\end{figure*}

\begin{figure*}[!t]
\centering
\includegraphics[width=\textwidth,height=0.88\textheight,keepaspectratio]{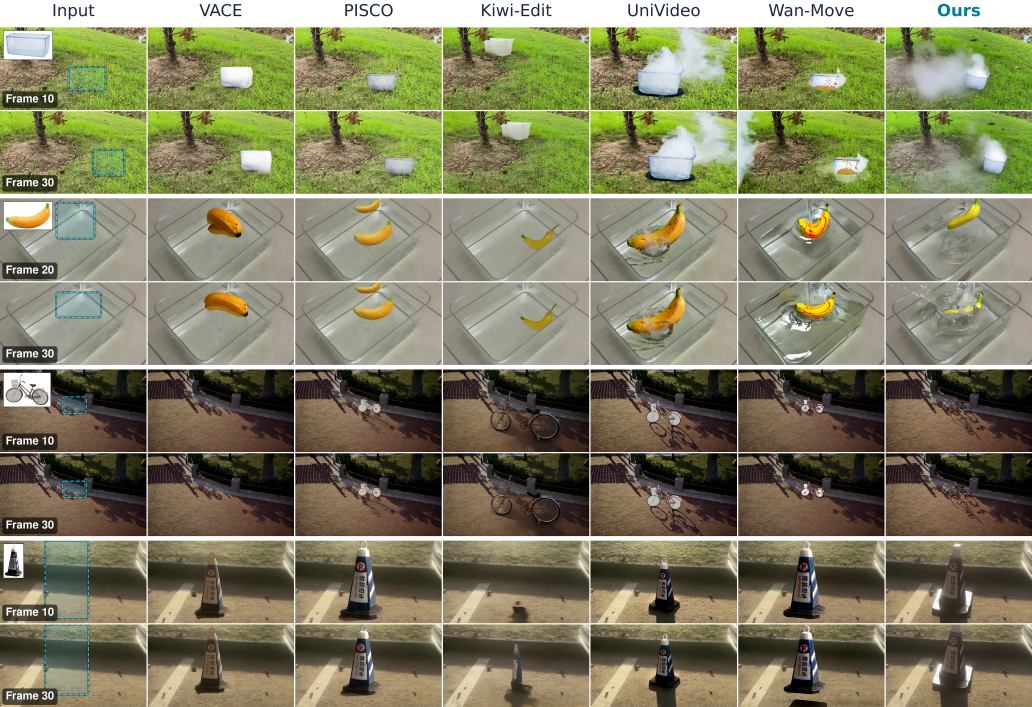}
\caption{\textbf{Qualitative comparison for video object insertion on VOR-Eval and ROSE-Bench.} Each case is shown at two frames (two rows); columns are methods. In the input, the cyan box marks where the object is inserted, the top-left inset shows the reference object, and the bottom-left label gives the frame index. Ours places the object in the box and synthesizes the effects of its interaction with the scene, such as the smoke of dry ice and the water splash, more faithfully than the baselines. \href{https://jiayi-wu-leo.github.io/duoedit/}{\textcolor{blue}{\textbf{More examples and full videos on the project page.}}}}
\label{fig:insertion_qualitative}
\end{figure*}

\begin{figure*}[!t]
\centering
\includegraphics[width=\textwidth]{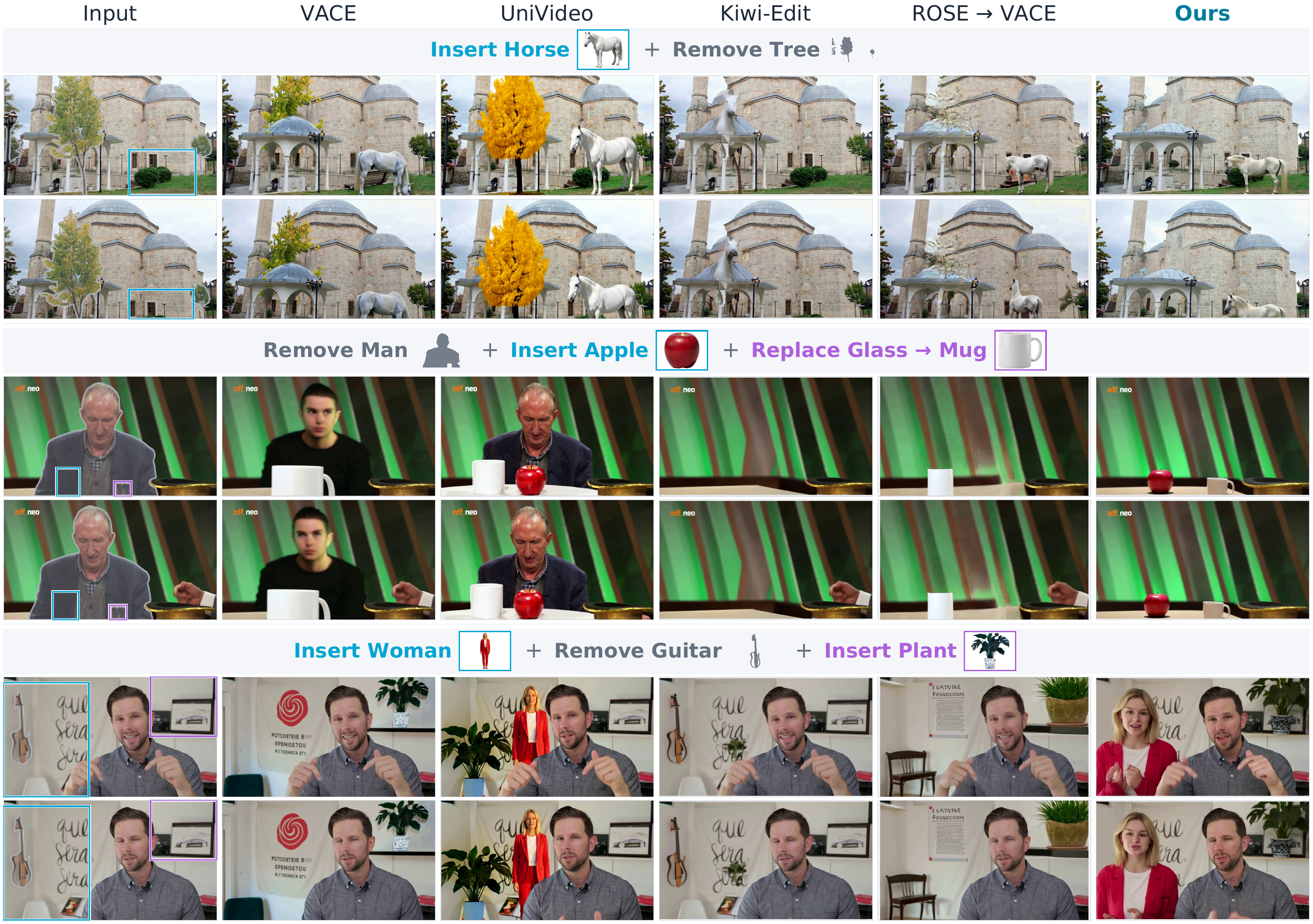}
\caption{\textbf{Compositional local video editing.}
Top: insert a horse and remove a tree.
Middle: remove a man, insert an apple, and replace a glass with a mug.
Bottom: insert a woman and a plant while removing a guitar.
Appearance references and spatial controls accompany each request; gray masks indicate removal, and matching colors link each reference to its target box.
Each example shows two corresponding frames across methods.
\method{} performs the requested operations jointly in a single inference run.
\href{https://jiayi-wu-leo.github.io/duoedit/}{\textcolor{blue}{\textbf{More examples and full videos on the project page.}}}}
\label{fig:compositional_qualitative}
\end{figure*}

\subsection{Associating Reference Appearance with Box Trajectories}
\label{sec:reference}

We associate each appearance reference with its corresponding box trajectory through a shared subject ID (\cref{fig:model_overview}).
Two reference canvases share the same layout: an appearance canvas contains the subject appearances, and an identity canvas encodes their IDs.
The box trajectories carry the same IDs, establishing correspondence between appearance references and spatial controls even when the reference layout differs from the target arrangement.

Reference and video tokens participate in shared self-attention, where the shared ID cues guide internal spatial--appearance routing.
This allows reference appearance and target placement to be specified independently while retaining their subject-level correspondence.

\subsection{Learning Two Primitives from Paired Videos}
\label{sec:primitive_training}

\method{} adapts a pretrained video diffusion Transformer (DiT)~\cite{wan22} to the unified editing interface.
Source-video latents, erasure-mask features, and box-trajectory latents are combined with the noisy target latent and projected into video tokens.
The paired reference canvases are encoded into reference tokens and prepended to the video token sequence.
Text supplies an additional conditioning signal to the shared transformer.

Training samples contain only insertion or removal, and both tasks share the same model and conditioning interface.
We optimize a flow-matching objective over the complete target video~\cite{lipman2022flow}.
To preserve pretrained generative priors while enabling parameter-efficient adaptation, we update only the input mapping layers for the new editing conditions and LoRA adapters~\cite{hu2021lora} within the DiT.
All other pretrained parameters, including those of the video autoencoder and text encoder, remain frozen.
At inference, all requested controls condition a single denoising process that produces the complete edited video.
Detailed network, optimization, and sampling settings are provided in \cref{sec:implementation}.

\section{Experiments}
\label{sec:experiments}

\subsection{Experimental Setup}
\label{sec:setup}

We fine-tune \method{} on VOR~\cite{fu2026effecterase} using the bidirectional data construction in Sec.~\ref{sec:bidirectional_data}, with supervision only for removal and insertion.
Training takes three days on two NVIDIA H200 GPUs, and we report results from the 20{,}000-step checkpoint.
Inference uses 25 denoising steps and a text guidance scale of 4.5.

We evaluate removal and insertion on the full VOR-Eval and ROSE-Bench~\cite{miao2025rose} benchmarks, which include object-related scene effects.
For each task, we report the mean of the two benchmark scores.
For qualitative removal comparisons, we additionally use VOR-Wild~\cite{fu2026effecterase}, the in-the-wild videos released with VOR, which have no object-absent ground truth.
Insertion examples are constructed by reversing removal pairs and preparing corresponding appearance references and spatial controls.
Replacement and mixed multi-subject, multi-operation editing are evaluated together on CoinVE-200K~\cite{long2026coinve}, without additional fine-tuning.
Each baseline receives the same source videos and editing requests through its native conditioning interface.
Implementation details and evaluation procedures are provided in Appendices~\ref{sec:implementation} and~\ref{sec:eval_details}.

\subsection{Video Object Removal}
\label{sec:removal_eval}

\paragraph{Protocol.}
Given a source video and removal masks, the task is to remove the designated subjects together with their associated scene effects.
We measure reconstruction against object-absent targets using full-frame PSNR, SSIM~\cite{wang2004ssim}, and LPIPS~\cite{zhang2018perceptual}, together with masked-region PSNR and LPIPS.
FVD~\cite{unterthiner2018towards}, VFID~\cite{wang2018video}, VLM-based scores, and VBench~\cite{huang2024vbench} provide complementary assessments of distributional quality, removal success, and video quality.
Following the VLM-as-a-judge paradigm~\cite{zheng2023judging}, VLM scores strictly follow the evaluation prompts of MiniMax-Remover~\cite{zi2026minimax}, with the open-source Qwen3.8-27B~\cite{qwen38} as the judge.

\begin{table}[!ht]
\centering
\caption{Object removal, mean of VOR-Eval and ROSE-Bench. Per column: {\setlength{\fboxsep}{1.5pt}\colorbox{red!30}{best}}, {\setlength{\fboxsep}{1.5pt}\colorbox{orange!35}{second best}}, {\setlength{\fboxsep}{1.5pt}\colorbox{yellow!45}{third best}}.}
\label{tab:obj_removal_mean}
\resizebox{\columnwidth}{!}{%
\begin{tabular}{l ccc cc cc cc c}\toprule
\multirow{2}{*}{\textbf{Method}} & \multicolumn{3}{c}{\textbf{Full frame vs.\ GT}} & \multicolumn{2}{c}{\textbf{Inside mask}} & \multicolumn{2}{c}{\textbf{Distribution}} & \multicolumn{2}{c}{\textbf{VLM judge}} & \textbf{VBench} \\
\cmidrule(lr){2-4}\cmidrule(lr){5-6}\cmidrule(lr){7-8}\cmidrule(lr){9-10}\cmidrule(lr){11-11}
 & PSNR $\uparrow$ & SSIM $\uparrow$ & LPIPS $\downarrow$ & PSNR $\uparrow$ & LPIPS $\downarrow$ & FVD $\downarrow$ & VFID $\downarrow$ & Succ $\uparrow$ & VQ $\uparrow$ & Overall $\uparrow$ \\
\midrule
\midrule
\textit{Input} & 21.45 & 0.826 & 0.191 & 11.17 & 0.716 & 380 & 0.721 & 0.04 & 1.69 & 0.8899 \\
\midrule
ProPainter~\cite{zhou2023propainter} & 24.10 & \cellcolor{orange!35}0.849 & 0.169 & 16.71 & 0.551 & 277 & 0.488 & 0.15 & 3.58 & 0.8807 \\
DiffuEraser~\cite{li2025diffueraser} & 24.68 & \cellcolor{red!30}0.853 & 0.162 & 18.09 & 0.448 & 250 & \cellcolor{yellow!45}0.420 & 0.30 & 4.62 & 0.8852 \\
Generative Omnimatte~\cite{lee2025generative} & 24.81 & 0.821 & 0.205 & 20.51 & 0.386 & \cellcolor{orange!35}176 & 0.542 & \cellcolor{yellow!45}0.68 & 6.08 & 0.8885 \\
MiniMax-Remover~\cite{zi2026minimax} & 24.01 & 0.833 & 0.166 & 18.00 & 0.425 & 222 & 0.457 & 0.51 & 5.04 & 0.8869 \\
ROSE~\cite{miao2025rose} & \cellcolor{red!30}27.12 & \cellcolor{yellow!45}0.840 & \cellcolor{red!30}0.143 & \cellcolor{orange!35}21.90 & \cellcolor{orange!35}0.309 & \cellcolor{orange!35}176 & \cellcolor{yellow!45}0.420 & 0.61 & \cellcolor{yellow!45}6.24 & 0.8904 \\
EffectErase~\cite{fu2026effecterase} & \cellcolor{yellow!45}25.08 & 0.821 & \cellcolor{yellow!45}0.158 & \cellcolor{yellow!45}20.80 & \cellcolor{yellow!45}0.329 & \cellcolor{yellow!45}184 & \cellcolor{orange!35}0.373 & \cellcolor{orange!35}0.77 & \cellcolor{orange!35}6.53 & \cellcolor{red!30}0.8947 \\
D2DF~\cite{chen2026draft} & 24.62 & 0.836 & 0.211 & 18.78 & 0.524 & 236 & 0.571 & 0.29 & 4.49 & 0.8784 \\
VACE~\cite{jiang2025vace} & 22.10 & 0.808 & 0.192 & 15.03 & 0.542 & 336 & 0.623 & 0.33 & 4.14 & \cellcolor{yellow!45}0.8907 \\
\midrule
\rowcolor{gray!10}
\textbf{Ours} & \cellcolor{orange!35}25.62 & 0.829 & \cellcolor{orange!35}0.144 & \cellcolor{red!30}22.05 & \cellcolor{red!30}0.277 & \cellcolor{red!30}109 & \cellcolor{red!30}0.346 & \cellcolor{red!30}0.85 & \cellcolor{red!30}6.63 & \cellcolor{orange!35}0.8911 \\
\bottomrule
\end{tabular}%
}
\end{table}

\paragraph{Results.}
Compared with dedicated state-of-the-art removal methods, \method{} achieves stronger distributional quality and removal success while maintaining competitive video quality (Table~\ref{tab:obj_removal_mean}).
Figure~\ref{fig:removal_qualitative} further demonstrates comparable or superior performance in jointly removing subjects and their associated scene effects, including mirror reflections, water ripples and wakes, illumination changes, and shadows.
These results show that learning removal within a unified editing framework retains the ability to handle the complex scene changes required by object removal.

\subsection{Video Object Insertion}
\label{sec:insertion_eval}

\paragraph{Protocol.}
Each insertion request consists of an object-absent source video, appearance references, and box trajectories.
We evaluate object placement using detection rate, mean box IoU, and normalized center distance.
Appearance fidelity is measured using CLIP~\cite{radford2021clip} and DINO~\cite{oquab2023dinov2} similarities to the corresponding ground-truth objects.
FVD~\cite{unterthiner2018towards}, VFID~\cite{wang2018video}, VLM-based scores, and VBench complement these measurements with assessments of realism, task fulfillment, and video quality.
Following the VLM-as-a-judge paradigm~\cite{zheng2023judging}, we adapt the evaluation prompts of OmniWeaving~\cite{pan2026omniweaving} for VLM scoring, with the open-source Qwen3.8-27B as the judge.

\begin{table}[!ht]
\centering
\caption{Object insertion, mean of VOR-Eval and ROSE-Bench. Per column: {\setlength{\fboxsep}{1.5pt}\colorbox{red!30}{best}}, {\setlength{\fboxsep}{1.5pt}\colorbox{orange!35}{second best}}, {\setlength{\fboxsep}{1.5pt}\colorbox{yellow!45}{third best}}.}
\label{tab:obj_insertion_mean}
\resizebox{\columnwidth}{!}{%
\begin{tabular}{l ccc cc cc cc c}\toprule
\multirow{2}{*}{\textbf{Method}} & \multicolumn{3}{c}{\textbf{Object placement}} & \multicolumn{2}{c}{\textbf{Fidelity to GT}} & \multicolumn{2}{c}{\textbf{Realism}} & \multicolumn{2}{c}{\textbf{VLM judge}} & \textbf{VBench} \\
\cmidrule(lr){2-4}\cmidrule(lr){5-6}\cmidrule(lr){7-8}\cmidrule(lr){9-10}\cmidrule(lr){11-11}
 & Det. $\uparrow$ & mIoU $\uparrow$ & CD $\downarrow$ & CLIP $\uparrow$ & DINO $\uparrow$ & FVD $\downarrow$ & VFID $\downarrow$ & IF $\uparrow$ & Avg $\uparrow$ & Overall $\uparrow$ \\ \midrule
\textit{Nothing inserted} & 0.477 & 0.063 & 0.204 & 0.797 & 0.534 & 377 & 0.725 & 1.00 & 1.16 & 0.7818 \\
\midrule
VACE~\cite{jiang2025vace} & 0.938 & 0.648 & 0.031 & \cellcolor{orange!35}0.854 & \cellcolor{yellow!45}0.714 & \cellcolor{orange!35}307 & \cellcolor{orange!35}0.647 & \cellcolor{yellow!45}3.57 & \cellcolor{red!30}3.55 & 0.8013 \\
PISCO~\cite{gao2026pisco} & 0.942 & \cellcolor{yellow!45}0.667 & \cellcolor{orange!35}0.025 & \cellcolor{yellow!45}0.839 & \cellcolor{orange!35}0.717 & 409 & 0.920 & \cellcolor{orange!35}3.58 & 3.49 & \cellcolor{yellow!45}0.8069 \\
Kiwi-Edit~\cite{lin2026kiwi} & 0.856 & 0.176 & 0.133 & 0.803 & 0.577 & \cellcolor{yellow!45}400 & \cellcolor{yellow!45}0.916 & 3.00 & 2.94 & 0.7913 \\
UniVideo~\cite{wei2025univideo} & \cellcolor{orange!35}0.966 & 0.328 & 0.086 & 0.833 & 0.646 & 513 & 1.097 & 3.51 & \cellcolor{orange!35}3.53 & \cellcolor{red!30}0.8275 \\
AnyV2V~\cite{ku2024anyv2v} & 0.934 & 0.559 & 0.038 & 0.806 & 0.657 & 839 & 3.416 & 3.46 & 3.22 & \cellcolor{orange!35}0.8074 \\
MagicMotion~\cite{li2025magicmotion} & \cellcolor{yellow!45}0.954 & \cellcolor{orange!35}0.676 & \cellcolor{yellow!45}0.026 & 0.827 & 0.678 & 566 & 1.583 & 3.52 & 3.35 & \cellcolor{orange!35}0.8074 \\
Wan-Move~\cite{chu2026wan} & 0.948 & \cellcolor{red!30}0.705 & \cellcolor{orange!35}0.025 & 0.808 & 0.632 & 567 & 1.074 & 3.30 & 3.00 & 0.7997 \\
\midrule\rowcolor{gray!10}
\textbf{Ours} & \cellcolor{red!30}0.969 & \cellcolor{yellow!45}0.667 & \cellcolor{red!30}0.024 & \cellcolor{red!30}0.887 & \cellcolor{red!30}0.757 & \cellcolor{red!30}278 & \cellcolor{red!30}0.600 & \cellcolor{red!30}3.62 & \cellcolor{yellow!45}3.51 & 0.7947 \\
\bottomrule\end{tabular}%
}
\end{table}

\paragraph{Results.}
\method{} combines accurate spatial control with stronger fidelity to the ground-truth subjects and better distributional quality than the compared insertion methods (Table~\ref{tab:obj_insertion_mean}).
Figure~\ref{fig:insertion_qualitative} further shows that \method{} places subjects at the specified locations while synthesizing their associated scene effects, including dry-ice smoke and water splashes, more faithfully to the paired targets.
These comparisons demonstrate its ability to coordinate subject appearance, placement, and effects within the surrounding scene.

\subsection{Compositional Local Video Editing}
\label{sec:interaction_eval}

\begin{table}[!t]
\centering
\caption{Compositional local video editing on CoinVE-200K: all 424 compositional editing clips and 941 instructions. SA: Semantic Accuracy; SPA: Scope Accuracy; EP: Edit Persistence; AN: Appearance Naturalness; SC: Scale Consistency; MN: Motion Naturalness; CP: Content Preservation; mIoU: Mean Intersection over Union. AC denotes the All-Correct rate: the percentage of clips passing every SA--MN check, excluding CP. All scores use a 0--100 scale ($\uparrow$). \colorbox{red!30}{Red}, \colorbox{orange!35}{orange}, and \colorbox{yellow!45}{yellow} indicate the top three results per column, respectively. Detailed metric definitions are provided in Appendix~\ref{sec:eval_details}.}
\label{tab:coinve_main_1col}
\footnotesize
\setlength{\tabcolsep}{2pt}
\renewcommand{\arraystretch}{1.08}
\resizebox{\columnwidth}{!}{%
\begin{tabular}{l ccc ccc c c c}
\toprule
& \multicolumn{3}{c}{\textbf{Edit. Acc.}} & \multicolumn{3}{c}{\textbf{Phys. Nat.}} & \textbf{Sem.} & \textbf{Spat.} & \textbf{Comp.} \\
\cmidrule(lr){2-4}\cmidrule(lr){5-7}\cmidrule(lr){8-8}\cmidrule(lr){9-9}\cmidrule(lr){10-10}
\textbf{Method} & SA & SPA & EP & AN & SC & MN & CP & mIoU & AC \\
\midrule
VACE~\cite{jiang2025vace} & 47.4 & 58.2 & 54.7 & 49.5 & 56.6 & 48.6 & \cellcolor{orange!35}69.2 & \cellcolor{yellow!45}29.5 & 3.1 \\
UniVideo~\cite{wei2025univideo} & \cellcolor{orange!35}67.3 & \cellcolor{yellow!45}59.7 & \cellcolor{orange!35}70.4 & \cellcolor{orange!35}83.6 & \cellcolor{orange!35}83.2 & \cellcolor{orange!35}71.7 & \cellcolor{yellow!45}60.2 & \cellcolor{orange!35}31.2 & \cellcolor{orange!35}11.6 \\
Kiwi-Edit~\cite{lin2026kiwi} & 50.0 & \cellcolor{red!30}72.2 & 50.4 & 51.4 & \cellcolor{yellow!45}61.8 & \cellcolor{yellow!45}51.2 & \cellcolor{red!30}73.8 & 18.5 & \cellcolor{yellow!45}7.3 \\
ROSE~\cite{miao2025rose}$\to$VACE~\cite{jiang2025vace} & \cellcolor{yellow!45}63.1 & 51.5 & \cellcolor{yellow!45}67.1 & \cellcolor{yellow!45}51.7 & 55.4 & 47.2 & 57.4 & 29.4 & 5.4 \\
\midrule
\rowcolor{gray!10}
\textbf{Ours} & \cellcolor{red!30}82.9 & \cellcolor{orange!35}63.4 & \cellcolor{red!30}82.9 & \cellcolor{red!30}84.4 & \cellcolor{red!30}84.7 & \cellcolor{red!30}78.8 & 59.5 & \cellcolor{red!30}61.2 & \cellcolor{red!30}19.8 \\
\bottomrule
\end{tabular}%
}
\end{table}

\paragraph{Protocol.}
We evaluate 941 compositional local editing instructions across all 424 compositional editing video clips in CoinVE-200K.
The seven CoinVE-Bench checklist metrics cover editing accuracy, physical naturalness, and content preservation, using 6{,}163 questions assessed by an MLLM judge.
We additionally report object-placement mIoU and All-Correct (AC), the percentage of clips passing every accuracy and naturalness check.
Comparisons include VACE~\cite{jiang2025vace}, UniVideo~\cite{wei2025univideo}, Kiwi-Edit~\cite{lin2026kiwi}, and the sequential ROSE$\to$VACE pipeline~\cite{miao2025rose,jiang2025vace}.

\paragraph{Results.}
\method{} achieves more reliable compositional editing than both unified video editors and the sequential ROSE$\to$VACE pipeline, with stronger joint edit fulfillment, spatial placement, and physical naturalness (Table~\ref{tab:coinve_main_1col}).
Figure~\ref{fig:compositional_qualitative} demonstrates how the two learned primitives support insertion, removal, and replacement across multiple subjects in a single inference run.

\subsection{Ablation Studies}
\label{sec:results}

We ablate two training-time design choices while keeping the remaining training configuration fixed.
All variants use the same inference pipeline, test cases, reference images, and random seeds.
Detailed configurations are provided in Appendix~\ref{sec:eval_details}.

\paragraph{Inpainting source conditioning (w/o full source).}
We replace the complete source video with the backbone's native inpainting input, masking out the editing region and retaining the corresponding pretrained source-input weights.
We evaluate this variant on removal.
With inpainting input, the model sees a hole instead of the subject to be removed, losing visual cues for harmonizing its associated scene effects.
As shown in Figure~\ref{fig:ablation_qualitative} (bottom), retaining the complete source video enables better joint removal of the subject and its associated effects.

\paragraph{Reference augmentation (w/o ref.\ aug.).}
We replace augmented, unaligned training references with the clearest masked object crops from the object-present videos.
Both models receive the same augmented references at test time, and we evaluate insertion.
As shown in Figure~\ref{fig:ablation_qualitative} (top), reference augmentation helps the model insert the reference subject at the specified location while adapting it to the scene.
Without augmentation, the model tends to copy and paste the reference image with little adaptation to the surroundings.

\par\addvspace{12pt plus 1fill}\noindent
\begin{minipage}{\columnwidth}
\centering
\includegraphics[width=\linewidth]{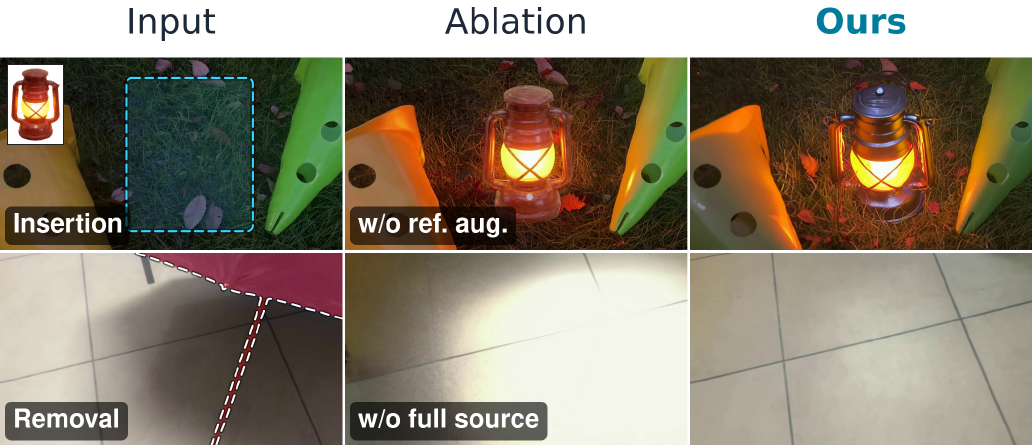}
\captionsetup{hypcap=false,skip=6pt}
\captionof{figure}{Qualitative ablation on VOR-Eval. Top: insertion; the input shows the target box (cyan) and the reference image (inset). Trained without reference augmentation, the model reproduces the reference image almost unchanged. Bottom: removal; the input shows the erasure mask in red. With inpainting source conditioning the model cannot see the removed umbrella and leaves an over-exposed region where its shadow was, while Ours removes the umbrella together with its shadow. \href{https://jiayi-wu-leo.github.io/duoedit/}{\textcolor{blue}{\textbf{More examples and full videos on the project page.}}}}
\label{fig:ablation_qualitative}
\end{minipage}
\par\addvspace{12pt plus 1fill}

\providecommand{\cmark}{\ding{51}}\providecommand{\xmark}{\ding{55}}

\par
\begin{samepage}
\section{Conclusion}
\label{sec:conclusion}

We presented \method{}, a unified local video editing model that learns insertion and removal as two independently controllable primitives. Composing their input conditions enables replacement and mixed-operation edits across multiple subjects in a single end-to-end inference run, without composition-specific training. Internal spatial--appearance routing associates unaligned appearance references with their corresponding box trajectories, providing separate control over appearance and placement. Both primitives are learned from the same object-removal dataset by reversing source--target pairs, requiring no additional paired training videos for compositional edits. This formulation supports flexible local editing while jointly considering subject interactions and harmonization with the surrounding scene.

\par
\end{samepage}

\clearpage
\endgroup

{
    \small
    \bibliographystyle{references}
    \bibliography{main}
}

\clearpage
\appendix
\begingroup
\color{blue}
\section*{Video Results}

We encourage readers to view the video results on our \href{https://jiayi-wu-leo.github.io/duoedit/}{\textcolor{blue}{project page}}, which include additional comparisons on compositional editing, object removal, object insertion, and ablation studies, as well as examples illustrating limitations.
\par
\endgroup

\section{Implementation Details}
\label{sec:implementation}

\subsection{Task Configuration}
\label{sec:task_implementation}

The unified interface represents a collection of edits without assigning one task type to the entire video.
Let $\mathcal D$ contain source subjects to remove or replace, and let $\mathcal A$ contain the appearances to insert, including replacements.
Their conditions are assembled into a shared erasure mask
\begin{equation}
    M_{\mathcal D}=\bigvee_{i\in\mathcal D}M_i,
    \label{eq:composed_mask}
\end{equation}
and a box video $B_{\mathcal A}$ and reference bank $R_{\mathcal A}$ for the synthesis set.
Pure removal contributes only to $\mathcal D$; insertion contributes only to $\mathcal A$; replacement contributes the old subject to $\mathcal D$ and its new appearance to $\mathcal A$.
The complete source, combined conditions, and a joint text instruction enter a single call to $G_{\theta}$.

\begin{table*}[t]
    \centering
    \small
    \setlength{\tabcolsep}{8pt}
    \begin{tabular}{llllll}
        \toprule
        Operation & Source & Erasure mask & Box trajectories & Reference bank & Adaptation \\
        \midrule
        Insertion & Object-absent $V^{-}$ & Empty & $B_A$ & $R_A$ & Yes \\
        Removal & Object-present $V^{+}$ & $M_A$ & Empty & Empty & Yes \\
        Replacement & Object-present $V^{+}$ & $M_A$ & $B_A$ & $R_B$ & No \\
        Mixed-operation edits & Source $S$ & $M_{\mathcal D}$ & $B_{\mathcal A}$ & $R_{\mathcal A}$ & No \\
        \bottomrule
    \end{tabular}
    \caption{Task configuration in \method{}.
    $A$ denotes the original subject set and $R_B$ provides the replacement appearances.
    The replacement row illustrates same-location replacement using the original box trajectories; in general, replacement trajectories specify the requested placement of the new subjects.
    Mixed-operation edits assign different operation types to different subjects; $\mathcal D$ and $\mathcal A$ are the erasure and synthesis sets defined above.
    An empty box or reference input is represented by a black image or video before encoding.
    The complete source video is supplied for every operation.}
    \label{tab:task_controls}
\end{table*}

Each example draw selects insertion with probability $0.5$ and removal otherwise.
The task draw varies with the example index and training epoch.
This supplies supervision in opposite directions on the same scene pair; it does not impose exact invertibility on the generated outputs.

For the same-location replacement configuration in Table~\ref{tab:task_controls}, the source remains $V^{+}$ and the erasure mask marks the original subjects.
We supply new appearances through $R_B$ while retaining the corresponding box trajectories $B_A$.
The controls are combined within one call to $G_{\theta}$, without generating an intermediate object-absent video.
No replacement pair is sampled during adaptation.
The held-out status of replacement refers to these adaptation tasks, rather than to the unknown coverage of the backbone's pretraining data.

\subsection{Source and Spatial Encoding}
\label{sec:source_conditioning}

\paragraph{Source visibility.}
We encode the complete source $S$ for every task, including removal and replacement.
The erasure mask never blacks out pixels in this source input.
Consequently, the denoiser can access the marked subject's appearance along with the surrounding scene.
This preserves evidence relevant to object-related effects that may extend beyond the silhouette.
The implementation denoises the entire target and performs no pixel-space paste-back outside $M$.

\paragraph{Spatial controls.}
The box video $B$ encodes subject trajectories using filled bounding boxes, with a distinct ID assigned to each subject.
Insertion and replacement use these box trajectories to specify placement over time; removal uses a black box video.
For removal training, we construct $M$ from the mask supplied with the paired data and dilate it by four pixels on the output grid.
The mask constructor falls back to box footprints if silhouettes are unavailable.
The source, target, boxes, and silhouettes use the same sampled frame indices.

\paragraph{Latent interface.}
Let $\mathcal{E}$ be the frozen video autoencoder.
Its video latents have 16 channels, with spatial compression by eight and temporal compression by four.
The binary mask is converted into four channels by repeating its first frame four times, grouping frames in fours, and resizing to the latent grid.
We denote this operation by $\mathcal{P}$.
The model retains the base mask polarity by encoding $1-M$.
For a noisy target latent $z_{\sigma}$, the input stack is
\begin{equation}
    u_{\sigma}=
    [z_{\sigma};\mathcal{E}(B);
    \mathcal{P}(1-M);\mathcal{E}(S)].
    \label{eq:latent_input}
\end{equation}
The channel counts are $16+16+4+16=52$, matching the pretrained input projection.
An empty pixel-space box input is still passed through $\mathcal{E}$; it is not defined to be a zero latent.

\subsection{Reference Canvas Construction}
\label{sec:reference_implementation}

The reference bank consists of two images with identical subject layouts.
The appearance canvas $R^{\mathrm{app}}$ contains the supplied appearance references on a black background.
During training, these are the augmented references constructed as described in \cref{sec:bidirectional_data}; at inference, the canvas packs the references supplied with the editing request.
We use the shared subject IDs described in \cref{sec:reference}.
The identity canvas $R^{\mathrm{id}}$ encodes each reference subject's ID within its silhouette.
The same ID is assigned to the subject's boxes in $B$, associating its appearance reference with the corresponding box trajectory.
Training samples a distinct visual encoding for each subject ID from a predefined set and shares this assignment between the box video and identity canvas.
The assignment changes across example and epoch draws.

The current implementation packs the subject crops while preserving their aspect ratios.
Relative tile sizes use the subjects' box sizes over the sampled clip, with layout augmentation.
Both canvases use exactly the same packing.
The association signal is therefore the shared subject ID, while the appearance canvas supplies the desired visual content.
Removal supplies two black canvases.

Each canvas is encoded as a single frame.
Their 16-channel latents are concatenated into
\begin{equation}
    r=
    [\mathcal{E}(R^{\mathrm{app}});
    \mathcal{E}(R^{\mathrm{id}})].
    \label{eq:reference_latent}
\end{equation}
A reference projection maps this 32-channel input into tokens prepended to the video token sequence.
The reference tokens occupy a leading position in the shared temporal positional grid and participate in the transformer's self-attention.
Their output positions are discarded before reconstructing the video latent.

\subsection{Backbone Adaptation and Optimization}
\label{sec:adaptation}

We use Wan2.2-Fun-A14B-Control~\cite{wan22}, retaining the high-noise and low-noise expert routing of Wan2.2~\cite{wan22}.
Each expert is adapted in a separate run over its assigned range of the shifted timestep schedule.
Both runs use the same two editing tasks and conditioning representation.
Inference selects the high-noise expert for timesteps at or above $900$ on the scheduler's 1000-step training scale, and the low-noise expert below this threshold.
The boundary is applied to timestep values, rather than to a fixed fraction of inference steps.

The video autoencoder, text encoder, and base transformer block weights remain frozen.
We insert LoRA~\cite{hu2021lora} adapters with rank 64 and scale parameter $\alpha=32$ into the query, key, value, and two feed-forward projections.
We also train the conditioning input projections.
The 52-channel video input projection retains pretrained weights for the box and mask channels.
Its source-video channel weights are initialized to zero because the source now remains visible under the mask.
The weights attached to the noisy target's 16 channels remain fixed.
The reference projection expands from 16 to 32 input channels, copying the existing appearance weights and initializing the added identity-channel weights to zero.

The adapters use learning rate $10^{-4}$ and weight decay $0.03$.
The input projections use learning rate $10^{-5}$ and no weight decay.
Their biases remain trainable.
For the video input projection, a gradient hook fixes the noisy-target channel weights; zero weight decay prevents an optimizer update to those fixed slices.

\subsection{Objective and Sampling}
\label{sec:objective}

For target latent $z_0=\mathcal{E}(Y)$ and Gaussian noise $\epsilon$, we use the linear conditional path
\begin{equation}
    z_{\sigma}=(1-\sigma)z_0+\sigma\epsilon,
    \qquad v^{*}=\epsilon-z_0.
    \label{eq:flow_path}
\end{equation}
The model predicts its conditional velocity following flow matching~\cite{lipman2022flow}.
Timestep indices are sampled within the expert's assigned range, using the base scheduler with shift 5.
Writing $d=v_{\theta}(z_{\sigma},\sigma;S,M,B,R,p)-v^{*}$, the implemented objective is
\begin{equation}
    \mathcal{L}
    =\mathbb{E}\!\left[
        \frac{1}{N}\sum_{j=1}^{N}
        d_j^2\,\mathbf{1}\{|d_j|\leq 50\}
    \right].
    \label{eq:loss}
\end{equation}
Here $N$ counts all target latent elements.
The elementwise gate excludes large residuals as a numerical safeguard.
The loss is averaged over the full video latent grid, with no additional edit-region weighting, cycle loss, or temporal-difference term enabled in the supplied configuration.

Insertion uses the scene's action and effect caption.
Removal samples a generic instruction describing removal of the marked subjects and their effects, without naming their identities.
The box, source-and-mask, and reference streams are independently dropped with probability $0.1$ during training.
Dropping the source stream sets its latent to zero and the latent mask to one.
These dropped states differ from encoding a black input for an empty task condition.
Text is additionally dropped after task assignment to support classifier-free text guidance.

Inference initializes the target latent from noise and integrates the predicted velocity with the flow scheduler.
The current inference configuration uses 25 steps and text guidance scale 4.5.
Its positive and negative text branches receive the same visual conditions.
The high- and low-noise experts use their respective adapted checkpoints, with the same task representation throughout the trajectory.

\section{Evaluation Protocol Details}
\label{sec:eval_details}

\subsection{Data and Sampling}

\paragraph{Training data.}
The VOR training set contains 49{,}626 video clips.

\paragraph{VOR-Eval.}
We evaluate removal and insertion on the full VOR-Eval benchmark, using the paired object-present and object-absent videos.
The paired videos support removal and insertion in opposite directions.
We use the first 81 consecutive frames, sampled at 16 fps.
The VOR training and evaluation manifests have no overlapping scene directory identifiers.

\paragraph{ROSE-Bench.}
We evaluate removal and insertion on all 60 paired clips in the ROSE-Bench~\cite{miao2025rose} benchmark manifest.
For removal, the object-present video is the input and the object-absent video is the target.
For insertion, we reverse the removal pairs and construct per-subject appearance references and spatial controls following \cref{sec:bidirectional_data}.
This insertion construction is our adaptation of the paired data.
For each task, we report the mean of the VOR-Eval and ROSE-Bench scores.

\paragraph{VOR-Wild.}
We use the in-the-wild videos released with VOR~\cite{fu2026effecterase} for additional qualitative removal comparisons.
These videos have no paired object-absent ground truth and are not included in the paired benchmark averages reported in Tables~\ref{tab:obj_removal_mean} and~\ref{tab:obj_insertion_mean}.

\paragraph{Inference settings.}
The configured VOR-Eval inference resolution preserves aspect ratio with a nominal pixel area of $960^2$.
\method{} uses 25 denoising steps, text guidance scale 4.5, and base random seed 42; its inference script offsets the seed by the scene index.
The high- and low-noise experts are evaluated as a checkpoint pair at 20{,}000 training steps.
Training takes three days on two NVIDIA H200 GPUs.

\subsection{Removal and Insertion Evaluation}

The measurements in this subsection apply to the removal and insertion benchmarks.
Measurements for compositional editing are specified in \cref{sec:coinve_eval_details}.

\paragraph{Paired reconstruction.}
For removal, full-frame PSNR, SSIM~\cite{wang2004ssim}, and LPIPS~\cite{zhang2018perceptual}, together with masked-region PSNR and LPIPS, compare an edited frame with the corresponding object-absent target frame.
Both videos use the same temporal window and spatial grid.
Using paired targets also permits evaluation of changes in shadows, reflections, and other object-related effects beyond the object mask.
For insertion, the object-present video supplies the ground truth for appearance and distributional comparisons below.
Replacement and mixed-operation editing are evaluated using the compositional protocol in \cref{sec:coinve_eval_details}.

\paragraph{Spatial control for insertion.}
All methods use a common object detector, object-query specification, score threshold, and assignment rule.
Detections are associated one-to-one with requested object tracks.
Detection rate and box IoU include all annotated visible object frames, assigning zero IoU to an unmatched object.
Center distance is normalized by the frame diagonal and reported for matched detections together with detection rate.

\paragraph{Appearance fidelity to ground truth.}
For insertion, CLIP-I~\cite{radford2021clip} and DINO-I~\cite{oquab2023dinov2} use cosine similarity between normalized image features of a generated object crop and the corresponding ground-truth object crop.
The DINO-I implementation uses DINOv2 features.
Crop construction must cover requested object regions independently of detector success, preventing failed generations from being silently excluded from the appearance comparison.
Object-frame scores are aggregated within each video before averaging across videos for the same task and dataset.

\paragraph{Distributional quality.}
We report FVD~\cite{unterthiner2018towards} and VFID~\cite{wang2018video} to compare generated videos with the corresponding ground-truth video distribution.
The object-absent videos provide the targets for removal, and the object-present videos provide the targets for insertion.

\paragraph{VLM-based evaluation.}
For removal, we follow the evaluation prompts of MiniMax-Remover~\cite{zi2026minimax}.
For insertion, we adapt the evaluation prompts of OmniWeaving~\cite{pan2026omniweaving}.
Both tasks use the open-source Qwen3.8-27B~\cite{qwen38} as the judge, following the VLM-as-a-judge paradigm~\cite{zheng2023judging}.

\paragraph{Video quality.}
We use VBench~\cite{huang2024vbench} background consistency, motion smoothness, temporal flickering, aesthetic quality, and imaging quality where applicable.
Subject consistency is evaluated for tasks that retain or synthesize a designated subject.
These scores characterize visual and temporal quality; the task-specific measurements establish whether the requested edit was performed.
Any confidence intervals are computed with videos as the sampling units.
For each task, we report the mean of the VOR-Eval and ROSE-Bench scores.

\subsection{Ablation Configurations}
\label{sec:ablation_details}

\paragraph{Controlled training and evaluation.}
All variants use the same training video pairs, two-expert training scheme with separate noise ranges, resolution, clip length, packed reference-canvas construction, LoRA configuration, optimizer settings, and number of training steps.
Evaluation uses an identical inference pipeline and the same test videos, editing requests, reference images, and random seeds.

\paragraph{Source-conditioning implementation.}
The full model receives the complete, unmasked source video together with a separate erasure mask and initializes the weights associated with the source-video input channels to zero.
The inpainting variant masks out the editing region and retains the corresponding pretrained weights, matching the backbone's pretraining input formulation.
As in the main-text ablation, we evaluate this variant on removal.
For insertion, the erasure mask is empty, so masking leaves the source-video input unchanged.

\paragraph{Reference construction.}
The full model uses training appearance references re-rendered with HunyuanImage-3.0-Instruct-Distil~\cite{cao2025hunyuanimage,tencent2026hunyuanimageinstructdistil} to vary viewpoint and illumination and complete occluded regions, producing references that are not pixel-aligned with the target.
The ablated model instead uses the clearest masked subject crop from the object-present video, retaining the viewpoint, illumination, and appearance of the corresponding target frame.
We evaluate this ablation on insertion, using the same augmented appearance references for both models.
This setting reflects practical reference images, such as product photographs or photographs captured elsewhere, whose pose and lighting typically differ from those of the target scene.

\subsection{Compositional Editing Evaluation}
\label{sec:coinve_eval_details}

We evaluate 941 compositional local editing instructions across all 424 compositional editing video clips in CoinVE-200K~\cite{long2026coinve}.

\paragraph{Checklist metrics.}
The CoinVE-Bench evaluation uses 6{,}163 questions assessed by an MLLM judge.
Editing accuracy comprises Semantic Accuracy (SA), which checks whether the requested operations are completed; Scope Accuracy (SPA), which checks whether changes are confined to the instructed scope; and Edit Persistence (EP), which checks whether edits remain consistent throughout the video without reappearance, ghosting, or disappearance.
Physical naturalness comprises Appearance Naturalness (AN), which assesses whether subject lighting, shadows, and material appearance fit the scene; Scale Consistency (SC), which assesses the plausibility of subject size; and Motion Naturalness (MN), which assesses natural motion without unintended drift.
Content Preservation (CP) evaluates the preservation of content outside the requested edits.
Its original 0--10 ratings are rescaled to 0--100 for reporting in Table~\ref{tab:coinve_main_1col}.
The six checklist metrics SA--MN are reported as percentages.

\paragraph{All-Correct rate.}
A clip is counted as All-Correct only if every question associated with SA, SPA, EP, AN, SC, and MN passes.
AC is the percentage of such clips among all evaluated clips.
CP is excluded from this criterion.

\paragraph{Spatial measurements.}
Spatial measurements apply only to added and replaced subjects.
We use OWLv2 to detect each subject and compare its predicted boxes with the corresponding ground-truth boxes frame by frame.
Mean IoU measures box overlap, assigning zero to missed detections.
Table~\ref{tab:coinve_main_1col} reports mIoU multiplied by 100, matching its 0--100 score scale.
Hit@0.5 measures the fraction of evaluated frames with IoU $\geq 0.5$, while Det\% measures the percentage of frames in which the subject is detected.
CDist measures the distance between predicted and ground-truth box centers, normalized by the frame diagonal, and is computed only for frames with detections.

\paragraph{Appearance and temporal consistency.}
For compositional editing, DINO-I and CLIP-I measure feature similarity between generated subject crops and their assigned reference images using DINOv2 and CLIP, respectively.
TempCons measures DINOv2 similarity between crops of the same subject in adjacent sampled frames, assessing temporal stability of its appearance.

\paragraph{Video quality.}
VBench evaluates visual and temporal video quality using its official normalization procedure, with a weight of 0.5 for the dynamic-degree dimension.

\end{document}